\documentclass{article}

\usepackage{arxiv}

\usepackage[utf8]{inputenc} 
\usepackage[T1]{fontenc}    
\usepackage{hyperref}       
\usepackage{url}            
\usepackage{booktabs}       
\usepackage{amsfonts}       
\usepackage{nicefrac}       
\usepackage{microtype}      
\usepackage{lipsum}		
\usepackage{graphicx}
\usepackage{natbib}
\usepackage{doi}

\usepackage{tabularx}
\usepackage{multirow}
\usepackage{comment}
\usepackage{amssymb}
\usepackage{amsmath}

\title{Foundational Study on Authorship Attribution of Japanese Web Reviews for Actor Analysis}

\author{ 
    Hiroshi Matsubara\textsuperscript{1},
    Shingo Matsugaya\textsuperscript{2,3},
    Taichi Aoki\textsuperscript{3},
    Masaki Hashimoto\textsuperscript{1} \\[6pt]
    \textsuperscript{1}Faculty of Engineering and Design, Kagawa University \\
    \textsuperscript{2}Trend Micro, Inc. \quad
    \textsuperscript{3}JC3 Japan Cybercrime Control Center \\[4pt]
    \texttt{\{s22t336, hashimoto.masaki\}@kagawa-u.ac.jp}
}

\renewcommand{\shorttitle}{Authorship Attribution of Japanese Web Reviews}

\hypersetup{
pdftitle={Foundational Study on Authorship Attribution of Japanese Web Reviews for Actor Analysis},
pdfauthor={Hiroshi Matsubara, Shingo Matsugaya, Taichi Aoki, Masaki Hashimoto},
pdfkeywords={authorship attribution, stylometry, threat intelligence, actor analysis, Japanese web reviews, BERT, top-k evaluation, dark web},
}

\begin{document}
\maketitle

\begin{abstract}
This study investigates the applicability of authorship attribution based on stylistic features to support actor analysis in threat intelligence.
As a foundational step toward future application to dark web forums, we conducted experiments using Japanese review data from clear web sources.
We constructed datasets from Rakuten Ichiba reviews and compared four methods: TF-IDF with logistic regression (TF-IDF+LR), BERT embeddings with logistic regression (BERT-Emb+LR), BERT fine-tuning (BERT-FT), and metric learning with $k$-nearest neighbors (Metric+kNN).
Results showed that BERT-FT achieved the best performance; however, training became unstable as the number of authors scaled to several hundred, where TF-IDF+LR proved superior in terms of accuracy, stability, and computational cost.
Furthermore, Top-$k$ evaluation demonstrated the utility of candidate screening, and error analysis revealed that boilerplate text, topic dependency, and short text length were primary factors causing misclassification.
\end{abstract}


\section{Introduction}
\label{sec:introduction}
 
In recent years, threat intelligence has increasingly emphasized the identification and tracking of threat actors behind cyberattacks.
Authorship attribution---the task of inferring the author of a text based on stylistic features---serves as one of the foundational technologies supporting this effort.
While dark web forum posts represent a valuable information source for actor analysis, significant constraints on data collection and sharing make it difficult to conduct reproducible evaluations.
 
Therefore, in this study, we first perform a foundational evaluation using reproducible Japanese review data from the clear web, with the aim of future application to dark web analysis.
Specifically, using the Rakuten Ichiba review dataset, we compare four methods under identical conditions: TF-IDF with logistic regression (TF-IDF+LR), BERT embeddings with logistic regression (BERT-Emb+LR), BERT fine-tuning (BERT-FT), and metric learning with $k$-nearest neighbors (Metric+kNN).
In addition to Accuracy and Macro-F1, we employ Top-$k$ evaluation to assess the utility of candidate screening.
We also analyze scaling characteristics with respect to the number of authors $U$, the number of posts per author $k$, and class imbalance conditions.
 
We note that this study is currently limited to Japanese reviews on the clear web and does not directly evaluate dark web posts, which involve additional challenges such as multilingual content, slang, short texts, and topic variability.
However, reviews share several difficulty factors with dark web posts---short text length, boilerplate expressions, and topic variability---making them suitable as a preliminary stage before isolating dark-web-specific factors.
By first conducting method comparison and error analysis under reproducible conditions, we can identify which conditions affect performance and what additional evaluation is needed for operational deployment.
 
The contributions of this study are as follows:
\begin{itemize}
    \item We compare representative authorship attribution methods under identical conditions on large-scale Japanese review data and organize the relationship between performance and computational cost.
    \item Through Top-$k$ evaluation, we verify the utility of candidate screening beyond strict identification.
    \item We organize the findings obtained under clear web, Japanese, and short-text constraints as challenges (limitations and additional evaluation items) for extension to the dark web.
\end{itemize}
 
The remainder of this paper is organized as follows.
Section~\ref{sec:related_work} reviews related work.
Section~\ref{sec:method} describes the dataset, preprocessing, methods, and evaluation.
Section~\ref{sec:results} presents experimental results and discussion.
Finally, Section~\ref{sec:conclusion} summarizes our findings and discusses future work.

\section{Related Work}
\label{sec:related_work}

Authorship attribution is a research field that infers the author of a text based on stylistic features.
Classical frameworks using features such as function words and vocabulary distributions have been studied since early work by \citet{mosteller1964inference}.
From the perspective of computational stylometry, feature design, classifiers, and evaluation settings have been systematically organized \citep{stamatatos2009survey}.
In recent years, methods using the pre-trained language model BERT \citep{devlin2019bertpretrainingdeepbidirectional} have become widespread, with studies reporting improved accuracy through fine-tuning \citep{fabien-etal-2020-bertaa}.
Table~\ref{tab:relatedwork} summarizes representative studies.
 
\subsection{Stylometry-Based Authorship Attribution}
\label{sec:rw_stylometry}
 
Classical authorship attribution typically employs features closer to writing habits than to content, such as function words and vocabulary distributions \citep{mosteller1964inference}.
\citet{stamatatos2009survey} organized authorship attribution research from the perspectives of features (character/word $n$-grams, etc.), classifiers (LR, SVM, etc.), and evaluation design.
In Japanese, author identification using stylistic features with machine learning has also been reported \citep{jin2007randomforest_ja}.
 
Meanwhile, scale conditions such as the number of authors $U$ and the number of posts per author $k$ have been shown to affect performance, with the problem becoming harder as data scale increases \citep{luyckx2010effect}.
Topic dependency in short texts \citep{stamatatos-2017-authorship} and handling domain differences \citep{kestemont2018overview} have also been identified as challenges.
 
\subsection{Deep Learning and Pre-trained Language Models (BERT)}
\label{sec:rw_bert}
 
With the advent of BERT \citep{devlin2019bertpretrainingdeepbidirectional}, authorship attribution leveraging contextual representations has become widespread.
A representative approach fine-tunes BERT for the authorship attribution task to improve accuracy \citep{fabien-etal-2020-bertaa}.
Additionally, approaches that learn author representations (embeddings) and perform identification based on distances (contrastive/distance learning) have been proposed \citep{huertas2022part}.
In Japanese, comparative analysis of BERT-based authorship attribution and performance improvement through ensemble methods have been reported \citep{2024bert_ensemble_ja}.
 
\subsection{Positioning of This Study}
\label{sec:rw_positioning}
 
Prior work has examined:
(i) systematization of feature design and classifiers based on computational stylometry \citep{stamatatos2009survey};
(ii) the impact of scale factors such as $U$ and $k$ on performance \citep{luyckx2010effect};
(iii) handling topic dependency and domain differences (e.g., text distortion and cross-domain evaluation) \citep{stamatatos-2017-authorship,kestemont2018overview}; and
(iv) accuracy improvements using BERT (fine-tuning and distance learning) \citep{fabien-etal-2020-bertaa,huertas2022part}
(see Table~\ref{tab:relatedwork}).
 
In contrast, this study is characterized by conducting a foundational evaluation using reproducible Japanese review data from the clear web, with the aim of extending to dark web post analysis (actor analysis) in the future.
Specifically, under the difficult conditions of Japanese and short text, we compare conventional methods (TF-IDF+LR), BERT-based methods (BERT-Emb+LR, BERT-FT), and distance learning (Metric Learning) using identical preprocessing and evaluation settings.
We organize:
(1) the impact of scale conditions ($U$, $k$) and imbalance on performance, stability, and computational cost; and
(2) the utility of candidate screening based on Top-$k$ evaluation.
Through this positioning, we aim to clarify which methods are effective under which conditions, not only for strict identification but also from the operational perspective of narrowing down candidates.

\begin{table*}[t]
  \centering
  \scriptsize
  \setlength{\tabcolsep}{3pt}
  \renewcommand{\arraystretch}{1.15}
  \begin{tabularx}{\textwidth}{@{}l l c c X c@{}}
    \toprule
    Author (Year) & Method & Lang. & Length & Dataset & Role \\
    \midrule
 
    \multicolumn{6}{@{}l}{\textbf{Classical Stylometry}}\\
    \addlinespace[2pt]
    Mosteller \& Wallace (1964) \citep{mosteller1964inference}
      & Bayes + function words
      & En & Long
      & Political documents, few authors
      & $\circ$ \\
    Stamatatos (2009) \citep{stamatatos2009survey}
      & Feature + classifier + evaluation (Survey)
      & Multi & --
      & Various corpora (survey)
      & $\circledast$ \\
    Jin \& Murakami (2007) \citep{jin2007randomforest_ja}
      & RF + stylistic features
      & Ja & Long
      & Novels/essays/diaries
      & $\circ$ \\
 
    \addlinespace[3pt]
    \multicolumn{6}{@{}l}{\textbf{Challenges of Conventional Methods (Scale, Topic Dependency)}}\\
    \addlinespace[2pt]
    Luyckx \& Daelemans (2010) \citep{luyckx2010effect}
      & Scale factor analysis ($U$, $k$)
      & En & Short
      & Email, etc. (variable $U$, $k$)
      & $\circledast$ \\
    Stamatatos (2017) \citep{stamatatos-2017-authorship}
      & Text distortion (topic mitigation)
      & Multi & --
      & Multiple conditions
      & $\circ$ \\
    Kestemont et al. (2018) \citep{kestemont2018overview}
      & Benchmark (PAN)
      & Multi & --
      & Multiple tasks (cross-domain, etc.)
      & $\circ$ \\
 
    \addlinespace[3pt]
    \multicolumn{6}{@{}l}{\textbf{Deep Learning / BERT (Comparison Axis of This Study)}}\\
    \addlinespace[2pt]
    Devlin et al. (2019) \citep{devlin2019bertpretrainingdeepbidirectional}
      & Pre-trained language model (BERT)
      & Multi & --
      & (Foundation model)
      & $\triangle$ \\
    Fabien et al. (2020) \citep{fabien-etal-2020-bertaa}
      & BERT fine-tuning (authorship attribution)
      & En & Med--Long
      & Email/blog, multiple datasets
      & $\circledast$ \\
    Huertas-Tato et al. (2022) \citep{huertas2022part}
      & Author repr. learning (contrastive/distance)
      & En & --
      & Multiple datasets
      & $\circledast$ \\
    Kanda \& Jin (2024) \citep{2024bert_ensemble_ja}
      & Japanese BERT comparison + ensemble
      & Ja & Long
      & Aozora Bunko (literature)
      & $\circ$ \\
 
    \textbf{This study}
      & \textbf{TF-IDF / BERT-Emb / BERT-FT / Metric}
      & \textbf{Ja} & \textbf{Short}
      & \textbf{Reviews (Rakuten Ichiba)}
      & \textbf{--} \\
 
    \bottomrule
  \end{tabularx}
  \caption{Comparison of related work (Short = short text, Long = long text, $\circledast$ = core, $\circ$ = supplementary, $\triangle$ = background).}
  \label{tab:relatedwork}
\end{table*}

\section{Dataset and Experimental Methods}
\label{sec:method}
 
\subsection{Dataset (Rakuten Ichiba Reviews)}
\label{sec:dataset}

The primary subject of this study is the Rakuten Ichiba dataset \citep{rakuten_market}, which is provided as a large-scale review dataset containing reviewer IDs.
 
Rakuten Ichiba is one of the largest e-commerce platforms in Japan, accumulating reviews across diverse product categories.
In this study, we treat the reviewer ID (a masked identifier) as the author ID and aggregate review texts by author to construct the evaluation dataset.
 
The provided data is in TSV format and includes reviewer ID (masked), review text, rating score, and registration date.
The target period is January 2015 through December 2019; this study primarily analyzes reviews from 2019.
 
For the 2019 reviews, the top 100 authors ($U=100$) account for a total of 68,222 reviews, while the top 1,000 authors ($U=1{,}000$) account for 294,443 reviews.
Detailed statistics are shown in Tables~\ref{tab:rakuten_stats} and \ref{tab:rakuten_stats_1000}.
 
The post count distribution of the top 100 authors exhibits an imbalanced structure, with a few prolific authors and the majority contributing a moderate number of posts.
This imbalance corresponds to the conditions evaluated in Experiment~2 (Section~\ref{subsec:exp2}).
Additionally, the character count per review has a median of 60 characters and a mean of 118.45 characters, indicating a predominantly short-text setting where stylistic features are less likely to manifest.

\begin{table}[t]
  \centering
  \caption{Rakuten Ichiba review dataset statistics (2019, top 100 authors).}
  \label{tab:rakuten_stats}
  \small
  \setlength{\tabcolsep}{4pt}
  \begin{tabular}{l r}
    \toprule
    Item & Value \\
    \midrule
    Number of authors $U$ & 100 \\
    Total reviews $N$ & 68,222 \\
    \midrule
    Posts/author (min--median--max) & 419 -- 548 -- 2,489 \\
    Posts/author (mean $\pm$ SD) & 682.22 $\pm$ 472.6 \\
    \midrule
    Chars/review (median) & 60 \\
    Chars/review (mean) & 118.45 \\
    Chars/review (95th percentile) & 432 \\
    \bottomrule
  \end{tabular}
\end{table}
 
\begin{table}[t]
  \centering
  \caption{Rakuten Ichiba review dataset statistics (2019, top 1,000 authors).}
  \label{tab:rakuten_stats_1000}
  \small
  \setlength{\tabcolsep}{4pt}
  \begin{tabular}{l r}
    \toprule
    Item & Value \\
    \midrule
    Number of authors $U$ & 1,000 \\
    Total reviews $N$ & 294,443 \\
    \midrule
    Posts/author (min--median--max) & 186 -- 242 -- 2,489 \\
    Posts/author (mean) & 294.44 \\
    \midrule
    Chars/review (median) & 53 \\
    Chars/review (mean) & 88.76 \\
    Chars/review (95th percentile) & 277 \\
    \bottomrule
  \end{tabular}
\end{table}

\subsection{Preprocessing}
\label{sec:preprocess}
 
To reduce noise in the review texts and unify model inputs, we applied the following preprocessing steps:
\begin{itemize}
  \item Removal of boilerplate patterns such as order numbers and URLs using regular expressions.
  \item Normalization of line breaks, tabs, and consecutive whitespace to reduce surface-level variation.
  \item Exclusion of extremely short texts (10 characters or fewer) with minimal informational content.
\end{itemize}
 
\subsection{Compared Methods}
\label{sec:methods}
 
We compare the following four methods under identical conditions:
\begin{itemize}
  \item \textbf{TF-IDF+LR}: Character $n$-gram TF-IDF features with logistic regression (lightweight baseline).
  \item \textbf{BERT-Emb+LR}: Fixed [CLS] embeddings from a pre-trained BERT model with logistic regression.
  \item \textbf{BERT-FT}: BERT fine-tuned for the authorship attribution task (sequence classification).
  \item \textbf{Metric+kNN}: Distance learning to learn author representations, followed by nearest-neighbor search for identification (naturally compatible with Top-$k$ evaluation).
\end{itemize}
 
\subsection{Experimental Environment and Training Settings}
\label{sec:training_setting}
 
\begin{itemize}
  \item OS: Windows 11 / Ubuntu 22.04 (WSL2)
  \item CPU: Intel Core Ultra 7 265K, RAM: DDR5 64\,GB
  \item GPU: NVIDIA GeForce RTX 5070 Ti 16\,GB
  \item Python: 3.10.12, PyTorch: 2.0.1, Transformers: 4.30.0
\end{itemize}

\begin{table*}[t]
  \centering
  \scriptsize
  \setlength{\tabcolsep}{3pt}
  \renewcommand{\arraystretch}{1.15}
  \begin{tabularx}{\textwidth}{@{}l X@{}}
    \toprule
    Item & Setting (Key Parameters) \\
    \midrule
    Common & Stratified 5-fold cross-validation (seed=42); Metrics: Accuracy, Macro-F1, Top-$k$ ($k\in\{3,5,10\}$) \\
    TF-IDF & Character 2-gram to 3-gram, max\_features=10,000 \\
    LR & $C=1.0$, solver=lbfgs, max\_iter=1,000 \\
    BERT (Emb/FT) & \texttt{cl-tohoku/bert-base-japanese-v3}, max length 128 \\
    BERT-FT & AdamW, epochs=10, batch=16, weight\_decay=0.01, warmup\_steps=100, fp16 \\
    Metric+kNN & BatchHardTripletLoss, normalize=True, kNN: cosine, $k=3$ \\
    \bottomrule
  \end{tabularx}
  \caption{Key training settings used in this study.}
  \label{tab:hyperparams}
\end{table*}

\subsection{Evaluation}
\label{sec:evaluation}
 
This study evaluates authorship attribution as a multi-class classification problem (closed-set), assessing both strict identification (Top-1) and the utility of candidate screening (Top-$k$).
Results are reported as means over stratified 5-fold cross-validation; Accuracy and Macro-F1 are reported as mean $\pm$ standard deviation, while Top-$k$ values are reported as means ($k\in\{3,5,10\}$).
Training and inference times are also recorded for practical considerations.
 
\subsubsection{Accuracy / Macro-F1}
 
Accuracy captures overall identification performance.
Since class-imbalanced conditions (varying post counts per author) can cause majority-class bias, we additionally report Macro-F1, which equally weights the F1 score of each class.
 
\subsubsection{Top-\texorpdfstring{$k$}{k} Accuracy (Candidate Screening Performance)}
 
Top-$k$ Accuracy measures the proportion of instances where the correct author appears among the top $k$ candidates ranked by model output scores.
In large-scale settings where exact identification may be difficult, this metric evaluates whether the system can narrow down candidates to a manageable set (screening capability).
 
\subsubsection{Computational Cost}
 
Training and inference times are measured per fold and used to compare methods and understand scaling trends with respect to $U$ and $k$.

\subsection{Experimental Design}
\label{sec:exp_design}
 
With practical deployment in mind, we conduct a staged evaluation across three experimental conditions:
 
\begin{itemize}
  \item \textbf{Experiment~1: Method Comparison ($U=100$, variable $k$)}.
  Using the top 100 authors from 2019 reviews, we randomly sample an equal number of $k\in\{100,200,300,400\}$ reviews per author and compare the four methods.
 
  \item \textbf{Experiment~2: Imbalance Effects ($U=100$, full data)}.
  Using all 68,222 reviews from the top 100 authors (ranging from 419 to 2,489 reviews per author), we evaluate performance under class imbalance.
  We also examine the effect of imposing a per-author maximum cap $K_{\max}$.
 
  \item \textbf{Experiment~3: Scaling Characteristics (variable $U$, $k=186$ fixed)}.
  We varied the number of authors among $U\in\{2, 5, 10, 20, 50, 100, 200, 400, 600, 800, 1{,}000\}$, sampling $k=186$ reviews per author, and analyzed performance degradation as the number of authors increases.
  Here, $k=186$ is the minimum number of reviews available when extending to 1,000 authors (i.e., the post count of the 1,000th-ranked author).
\end{itemize}
 
\begin{figure}[t]
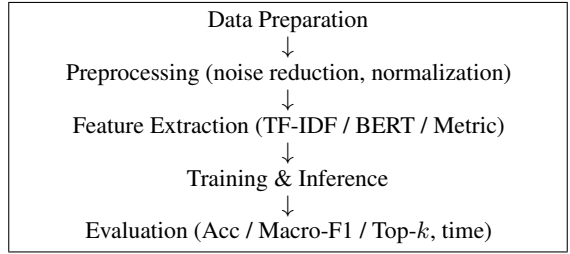

\centering
\fbox{\parbox{7.2cm}{\centering
\small
Data Preparation\\
$\downarrow$\\
Preprocessing (noise reduction, normalization)\\
$\downarrow$\\
Feature Extraction (TF-IDF / BERT / Metric)\\
$\downarrow$\\
Training \& Inference\\
$\downarrow$\\
Evaluation (Acc / Macro-F1 / Top-$k$, time)
}}
\caption{Overview of the evaluation pipeline.}
\label{fig:overview}
\end{figure}

Figure~\ref{fig:overview} illustrates the overall evaluation pipeline.
With preprocessing and evaluation conditions (stratified 5-fold) held constant, we vary the feature extraction and learning methods to compare performance and computational cost.

\section{Experimental Results and Discussion}
\label{sec:results}
 
\subsection{Experiment~1: Method Comparison (\texorpdfstring{$U=100$}{U=100}, Variable \texorpdfstring{$k$}{k})}
\label{subsec:exp1}
 
Using the top 100 authors ($U=100$), we sampled an equal number of $k\in\{100,200,300,400\}$ reviews per author and compared the four methods under identical conditions.
Table~\ref{tab:exp1_summary} shows results across all conditions.

\begin{table*}[t]
  \centering
  \scriptsize
  \setlength{\tabcolsep}{3pt}
  \renewcommand{\arraystretch}{1.15}
  \caption{Experiment~1 ($U=100$): Method comparison ($k\in\{100,200,300,400\}$, 5-fold mean).}
  \label{tab:exp1_summary}
  \begin{tabular}{lcccccccc}
    \toprule
    & \multicolumn{2}{c}{$k=100$} & \multicolumn{2}{c}{$k=200$} & \multicolumn{2}{c}{$k=300$} & \multicolumn{2}{c}{$k=400$} \\
    \cmidrule(lr){2-3}\cmidrule(lr){4-5}\cmidrule(lr){6-7}\cmidrule(lr){8-9}
    Method & Acc & Macro-F1 & Acc & Macro-F1 & Acc & Macro-F1 & Acc & Macro-F1 \\
    \midrule
    TF-IDF+LR   & 0.7372 & 0.7343 & 0.7903 & 0.7889 & 0.8157 & 0.8156 & 0.8314 & 0.8323 \\
    BERT-Emb+LR & 0.7059 & 0.7059 & 0.7467 & 0.7464 & 0.7636 & 0.7644 & 0.7772 & 0.7781 \\
    BERT-FT     & 0.7663 & 0.7662 & 0.8143 & 0.8127 & 0.8420 & 0.8415 & 0.8624 & 0.8597 \\
    Metric+kNN  & 0.6902 & 0.6778 & 0.7369 & 0.7283 & 0.7595 & 0.7542 & 0.7793 & 0.7735 \\
    \midrule
    & \multicolumn{2}{c}{$k=100$} & \multicolumn{2}{c}{$k=200$} & \multicolumn{2}{c}{$k=300$} & \multicolumn{2}{c}{$k=400$} \\
    \cmidrule(lr){2-3}\cmidrule(lr){4-5}\cmidrule(lr){6-7}\cmidrule(lr){8-9}
    Method & Top-5 & Top-10 & Top-5 & Top-10 & Top-5 & Top-10 & Top-5 & Top-10 \\
    \midrule
    TF-IDF+LR   & 0.8910 & 0.9292 & 0.9176 & 0.9504 & 0.9326 & 0.9603 & 0.9414 & 0.9655 \\
    BERT-Emb+LR & 0.8851 & 0.9337 & 0.9082 & 0.9518 & 0.9197 & 0.9580 & 0.9252 & 0.9607 \\
    BERT-FT     & 0.9045 & 0.9419 & 0.9295 & 0.9568 & 0.9430 & 0.9640 & 0.9494 & 0.9671 \\
    Metric+kNN  & 0.7915 & 0.8076 & 0.8233 & 0.8382 & 0.8433 & 0.8567 & 0.8551 & 0.8680 \\
    \bottomrule
  \end{tabular}
\end{table*}

\begin{figure}[t]
  \centering
  \includegraphics[width=\linewidth]{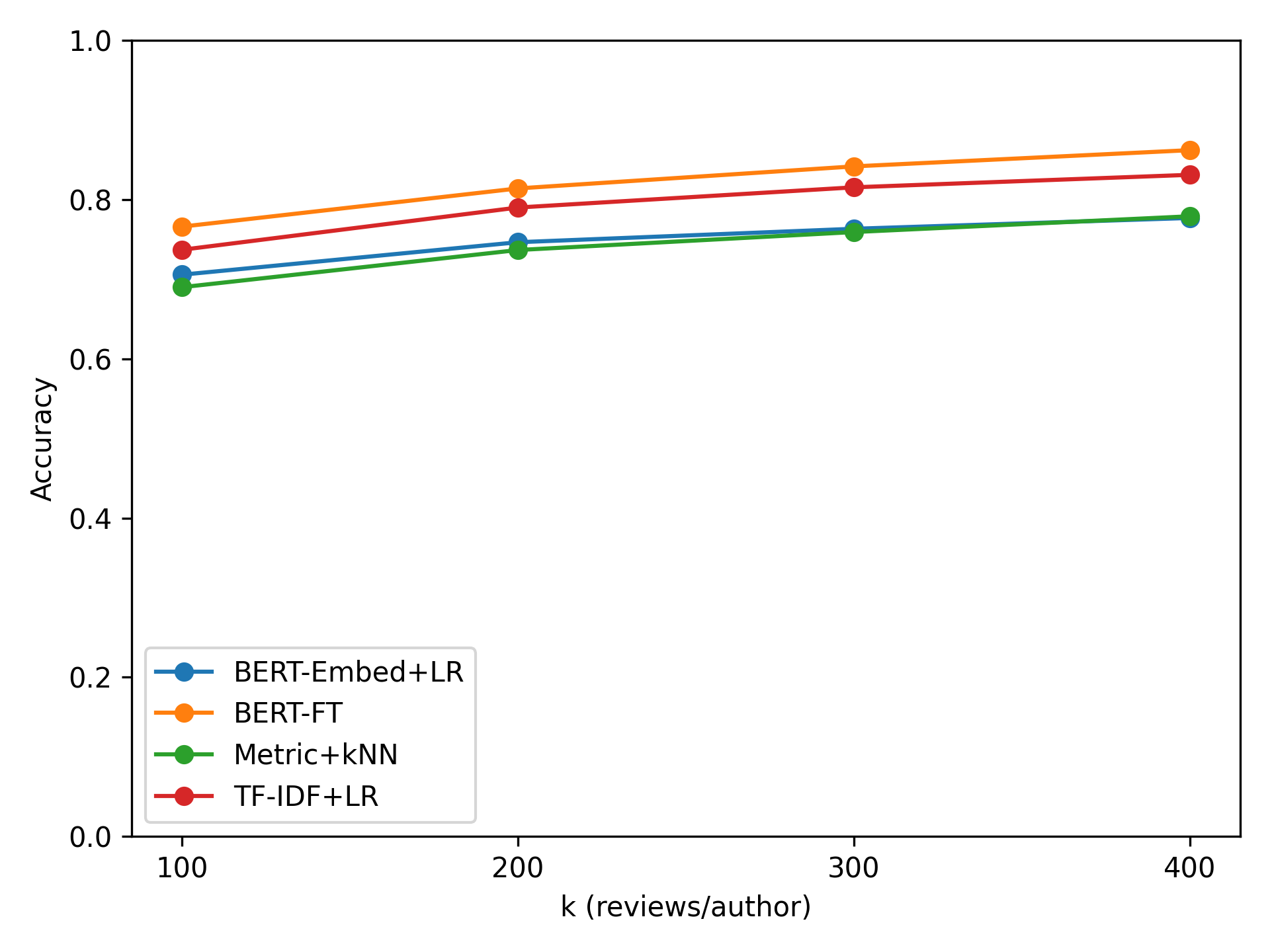}
  \caption{Experiment~1: Relationship between reviews per author $k$ and Accuracy ($U=100$).}
  \label{fig:exp1_accuracy_f1}
\end{figure}
 
\begin{figure}[t]
  \centering
  \includegraphics[width=\linewidth]{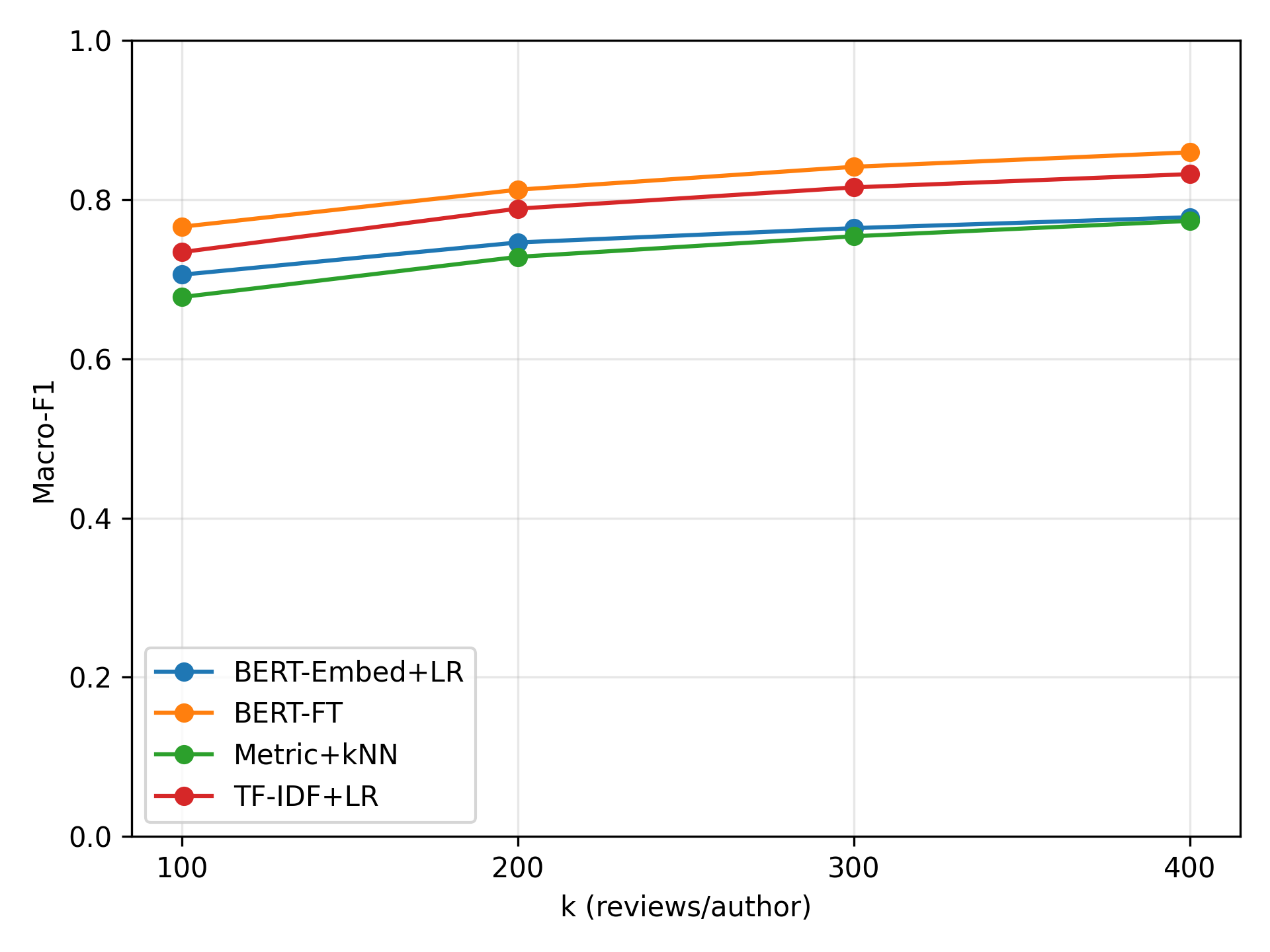}
  \caption{Experiment~1: Relationship between reviews per author $k$ and Macro-F1 ($U=100$).}
  \label{fig:exp1_macro_f1}
\end{figure}
 
\begin{figure*}[t]
  \centering
  \begin{minipage}{0.32\textwidth}
    \centering
    \includegraphics[width=\linewidth]{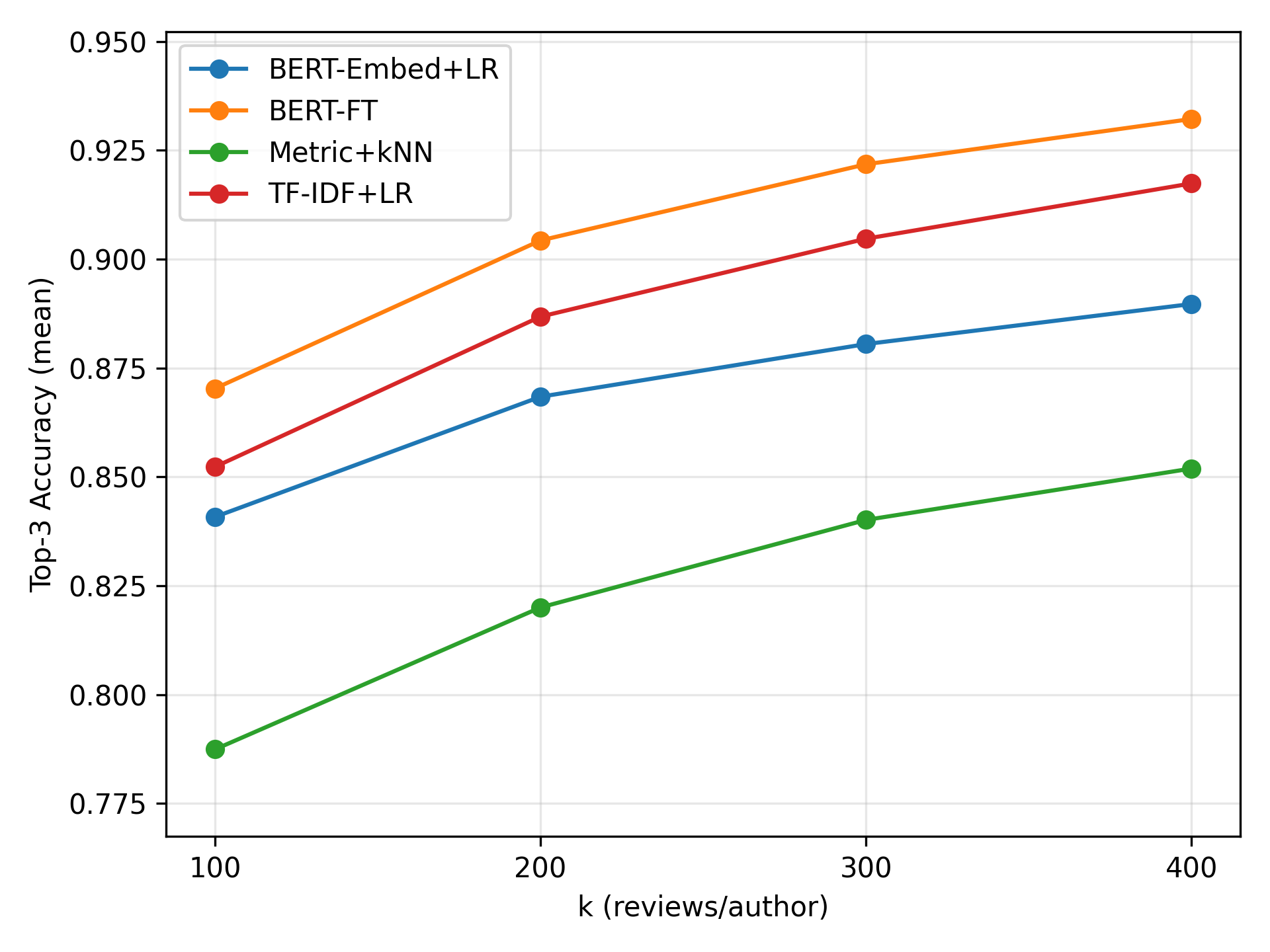}\\
    {\footnotesize (a) Top-3}
  \end{minipage}\hfill
  \begin{minipage}{0.32\textwidth}
    \centering
    \includegraphics[width=\linewidth]{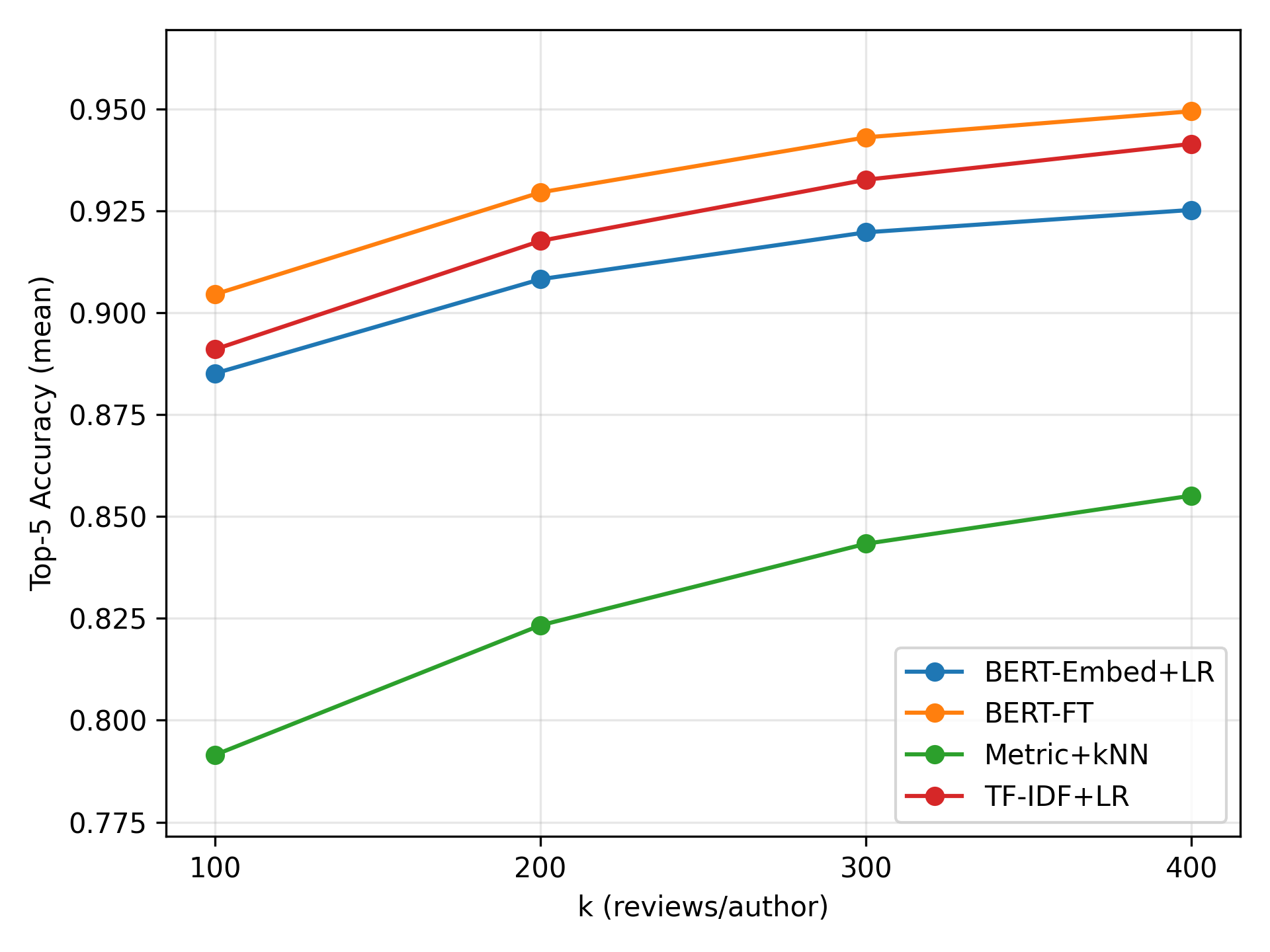}\\
    {\footnotesize (b) Top-5}
  \end{minipage}\hfill
  \begin{minipage}{0.32\textwidth}
    \centering
    \includegraphics[width=\linewidth]{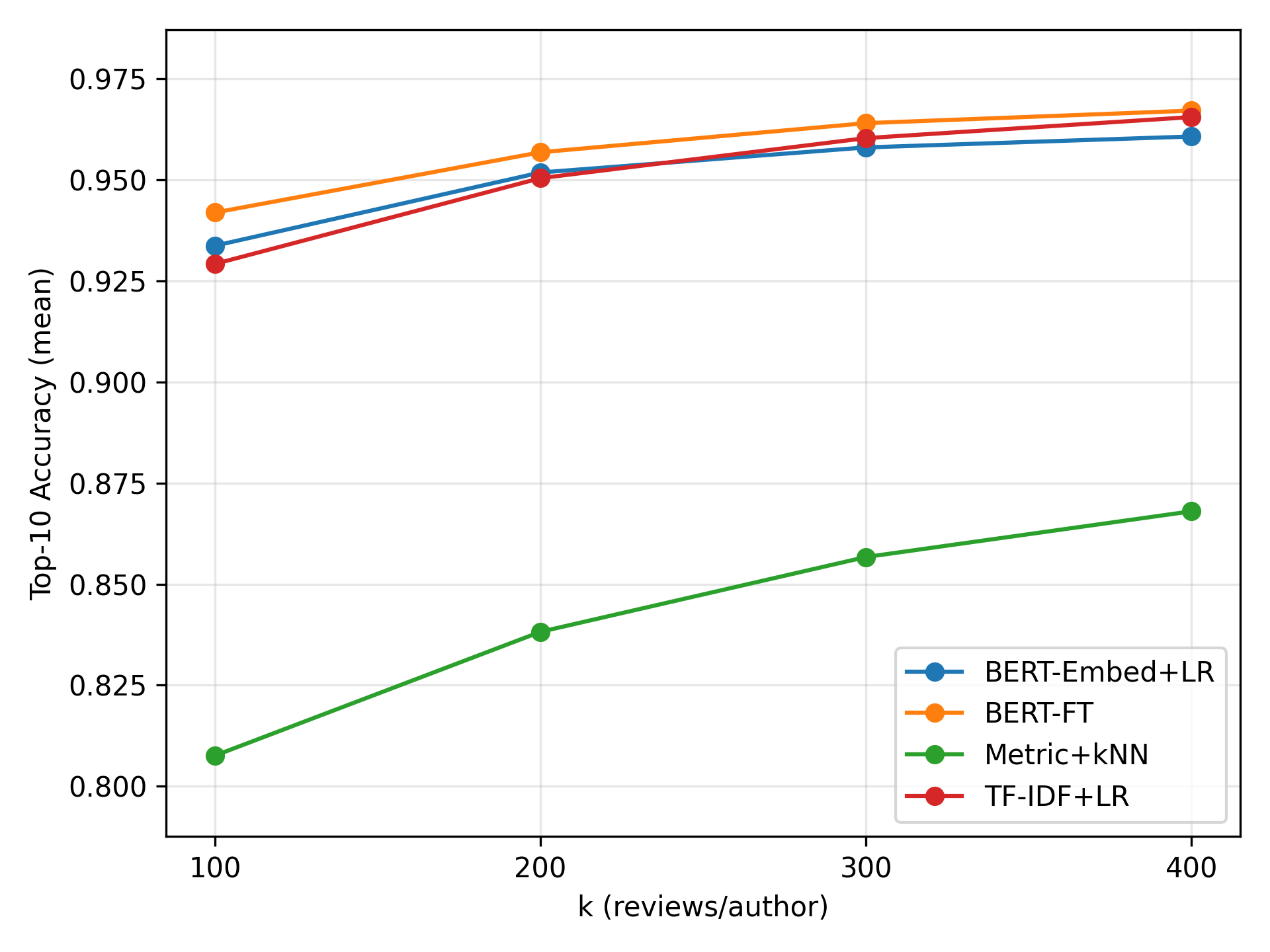}\\
    {\footnotesize (c) Top-10}
  \end{minipage}
  \caption{Relationship between $k$ and Top-$k$ performance at $U=100$.}
  \label{fig:exp1_k_trend_all}
\end{figure*}

\noindent\textbf{Results.}
BERT-FT achieved the highest performance across all values of $k$, reaching Accuracy 0.86 and Macro-F1 0.86 at $k=400$.
TF-IDF+LR was the second-best method with Accuracy 0.83 at $k=400$, while BERT-Emb+LR fell below TF-IDF+LR.
Metric+kNN exhibited the lowest performance across all conditions, with Top-5 also lower than other methods (0.855 at $k=400$).
Performance improved for all methods as $k$ increased, though the rate of improvement showed diminishing returns (Figure~\ref{fig:exp1_macro_f1}).
 
\noindent\textbf{Discussion.}
The superiority of BERT-FT is attributed to its context-aware representations and task-specific fine-tuning.
The underperformance of BERT-Emb+LR relative to TF-IDF+LR suggests that fixed embeddings from pre-trained BERT are insufficient for separating authorial style differences.
The low performance of Metric+kNN can be attributed to several factors:
(a) information constraints of kNN with $k=3$;
(b) reduced training efficiency due to batch size 16 being unable to cover all 100 authors; and
(c) insufficiently discriminative embeddings under short, boilerplate-heavy texts.

\subsection{Experiment~2: Effects of Post Count Distribution (Imbalance / \texorpdfstring{$K_{\max}$}{Kmax} Cap)}
\label{subsec:exp2}

Because BERT-Emb+LR and Metric+kNN were consistently inferior in Experiment~1, the subsequent experiments focus on the two strongest methods: TF-IDF+LR and BERT-FT.
 
Experiment~2A uses all data from the top 100 authors (imbalanced, 68,222 total reviews).
Experiment~2B imposes a per-author maximum post cap $K_{\max}$ and examines its effect on performance.
 
\noindent\textbf{Results.}
Under the imbalanced condition (Experiment~2A), BERT-FT achieved Accuracy 0.883 and TF-IDF+LR achieved 0.862, both higher than their Experiment~1 results at $k=400$ (Table~\ref{tab:exp2A_summary}).
However, training time differed substantially: 1,064 seconds/fold for BERT-FT versus 37 seconds/fold for TF-IDF+LR.
Under the $K_{\max}$ constraint (Experiment~2B), performance improved for both methods as the cap increased, with Top-5 remaining above 0.95 across all conditions (Table~\ref{tab:exp2B_summary}).

\begin{table}[t]
  \centering
  \small
  \setlength{\tabcolsep}{2pt}
  \renewcommand{\arraystretch}{1.15}
  \caption{Experiment~2A: Performance and computational cost under imbalance (full data, $U=100$, 5-fold mean; time in seconds/fold).}
  \label{tab:exp2A_summary}
  \begin{tabular}{lccc rr}
    \toprule
    Method & Acc & Macro-F1 & Top-5 & Train & Infer \\
    \midrule
    TF-IDF+LR & 0.8622 & 0.8459 & 0.9576 & 36.72  & 0.94 \\
    BERT-FT   & 0.8827 & 0.8610 & 0.9621 & 1063.99 & 7.70 \\
    \bottomrule
  \end{tabular}
\end{table}
 
\begin{table}[t]
  \centering
  \scriptsize
  \setlength{\tabcolsep}{3pt}
  \renewcommand{\arraystretch}{1.15}
  \caption{Experiment~2B: Effect of $K_{\max}$ cap ($U=100$, 5-fold mean).}
  \label{tab:exp2B_summary}
  \begin{tabular}{rccc ccc}
    \toprule
    $K_{\max}$ & \multicolumn{3}{c}{TF-IDF+LR} & \multicolumn{3}{c}{BERT-FT} \\
    \cmidrule(lr){2-4}\cmidrule(lr){5-7}
     & Acc & Macro-F1 & Top-5 & Acc & Macro-F1 & Top-5 \\
    \midrule
    500  & 0.8426 & 0.8416 & 0.9484 & 0.8489 & 0.8454 & 0.9510 \\
    1000 & 0.8567 & 0.8473 & 0.9548 & 0.8694 & 0.8569 & 0.9579 \\
    1500 & 0.8584 & 0.8464 & 0.9561 & 0.8710 & 0.8537 & 0.9592 \\
    \bottomrule
  \end{tabular}
\end{table}

\noindent\textbf{Discussion.}
While BERT-FT showed improved performance under imbalanced conditions, its training cost was approximately 29 times that of TF-IDF+LR, which can be a constraint for repeated application in operational settings.
The trend of performance improvement with increasing $K_{\max}$ indicates that additional data contributes to learning stylistic patterns; however, the improvement plateaus beyond $K_{\max}\ge1{,}000$, suggesting that simply adding more data eventually yields diminishing returns.

\subsection{Experiment~3: Scaling Characteristics (Variable \texorpdfstring{$U$}{U}, \texorpdfstring{$k=186$}{k=186} Fixed)}
\label{subsec:exp3}
 
We varied the number of authors among $U\in\{2, 5, 10, 20, 50, 100, 200, 400, 600, 800, 1{,}000\}$ with $k=186$ reviews sampled per author, and evaluated scaling behavior.
Here, $k=186$ is the minimum post count available when extending to 1,000 authors.
 
\begin{figure*}[t]
  \centering
  \begin{minipage}{0.49\textwidth}
    \centering
    \includegraphics[width=\linewidth]{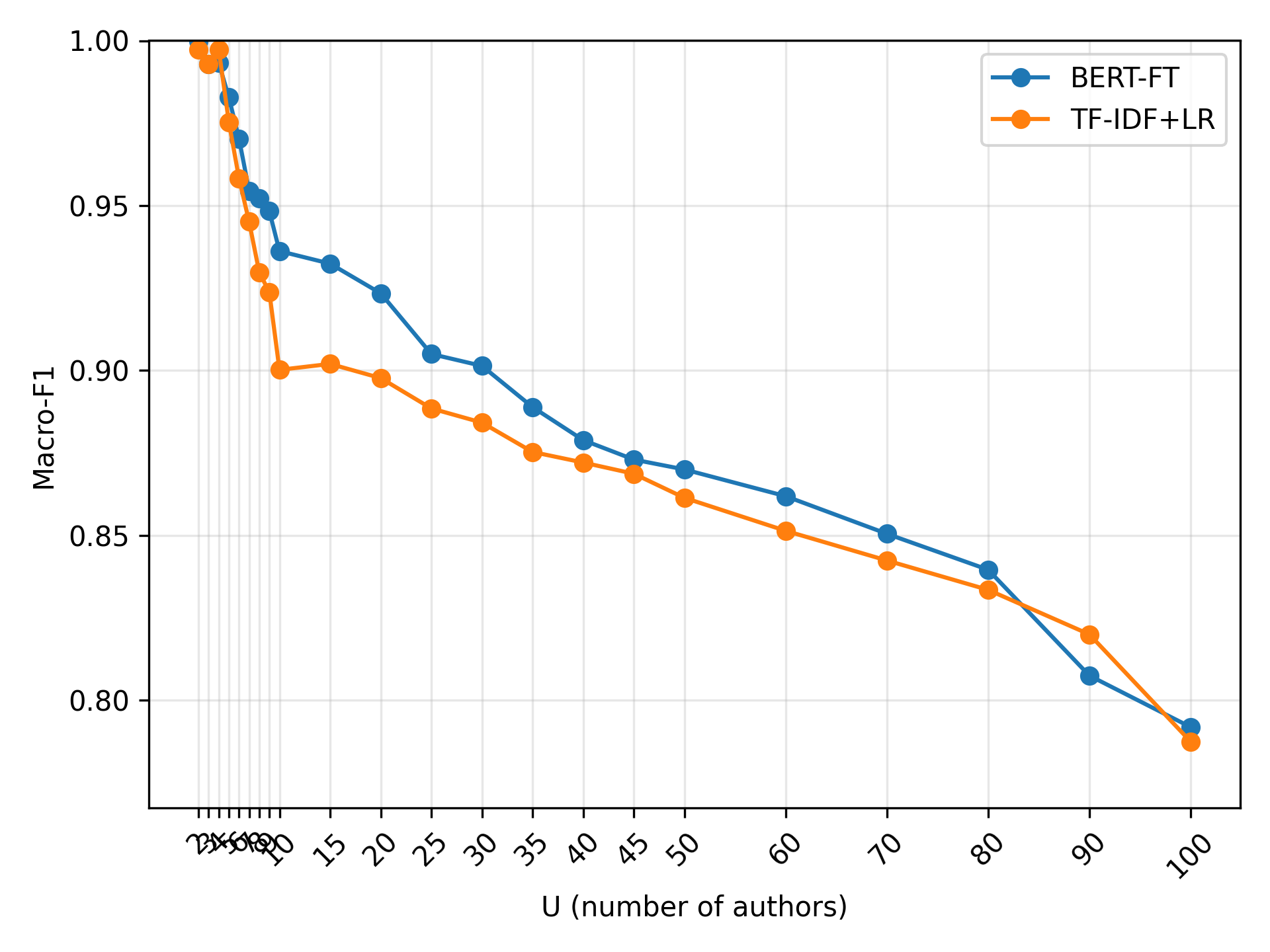}
    \caption{Experiment~3 (small scale): $U=2$--$100$.}
    \label{fig:exp3_smallU_macro_f1}
  \end{minipage}\hfill
  \begin{minipage}{0.49\textwidth}
    \centering
    \includegraphics[width=\linewidth]{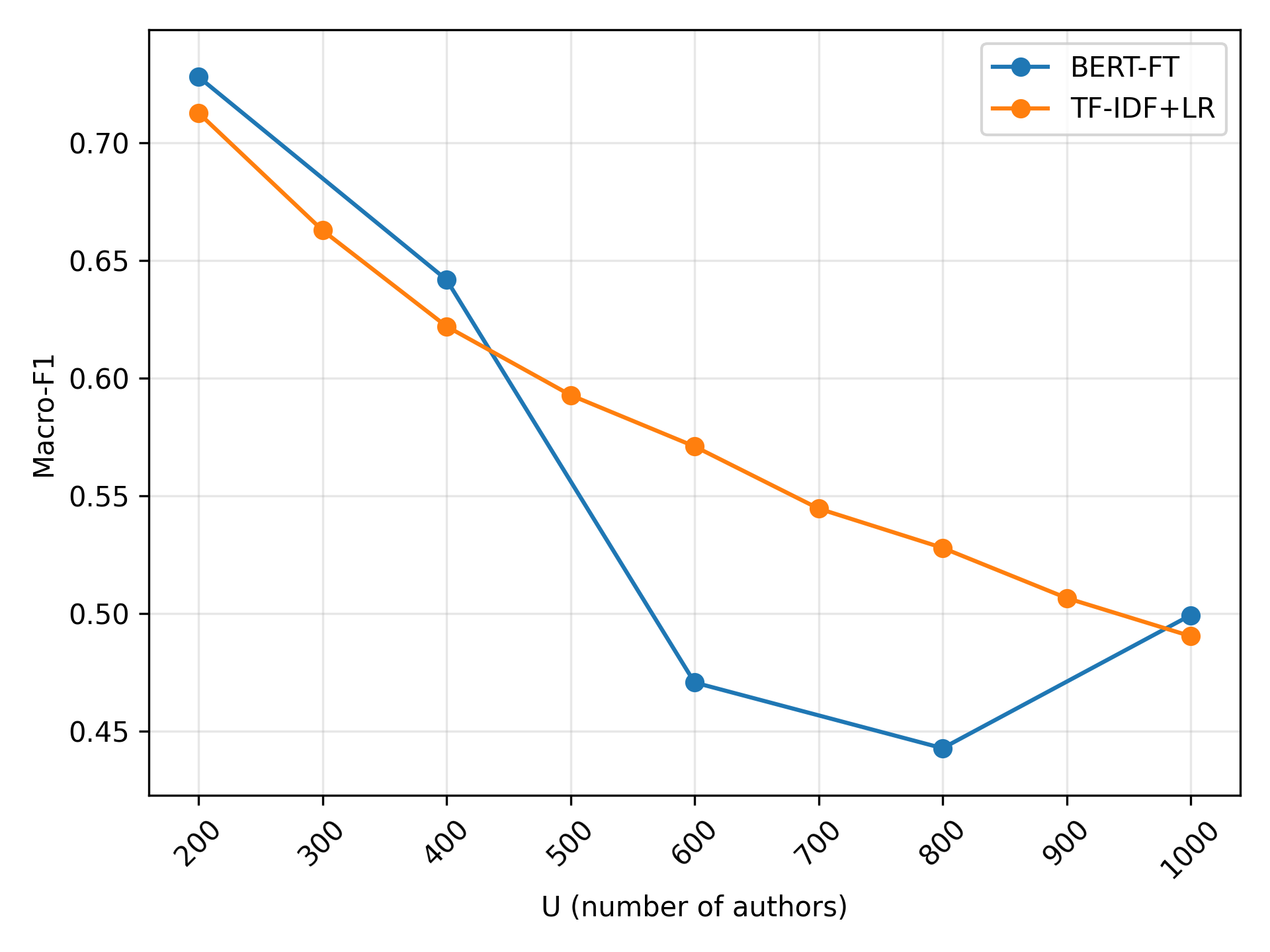}
    \caption{Experiment~3 (large scale): $U=200$--$1{,}000$.}
    \label{fig:exp3_largeU_macro_f1}
  \end{minipage}
\end{figure*}

\begin{figure*}[t]
  \centering
  \begin{minipage}{0.49\textwidth}
    \centering
    \includegraphics[width=\linewidth]{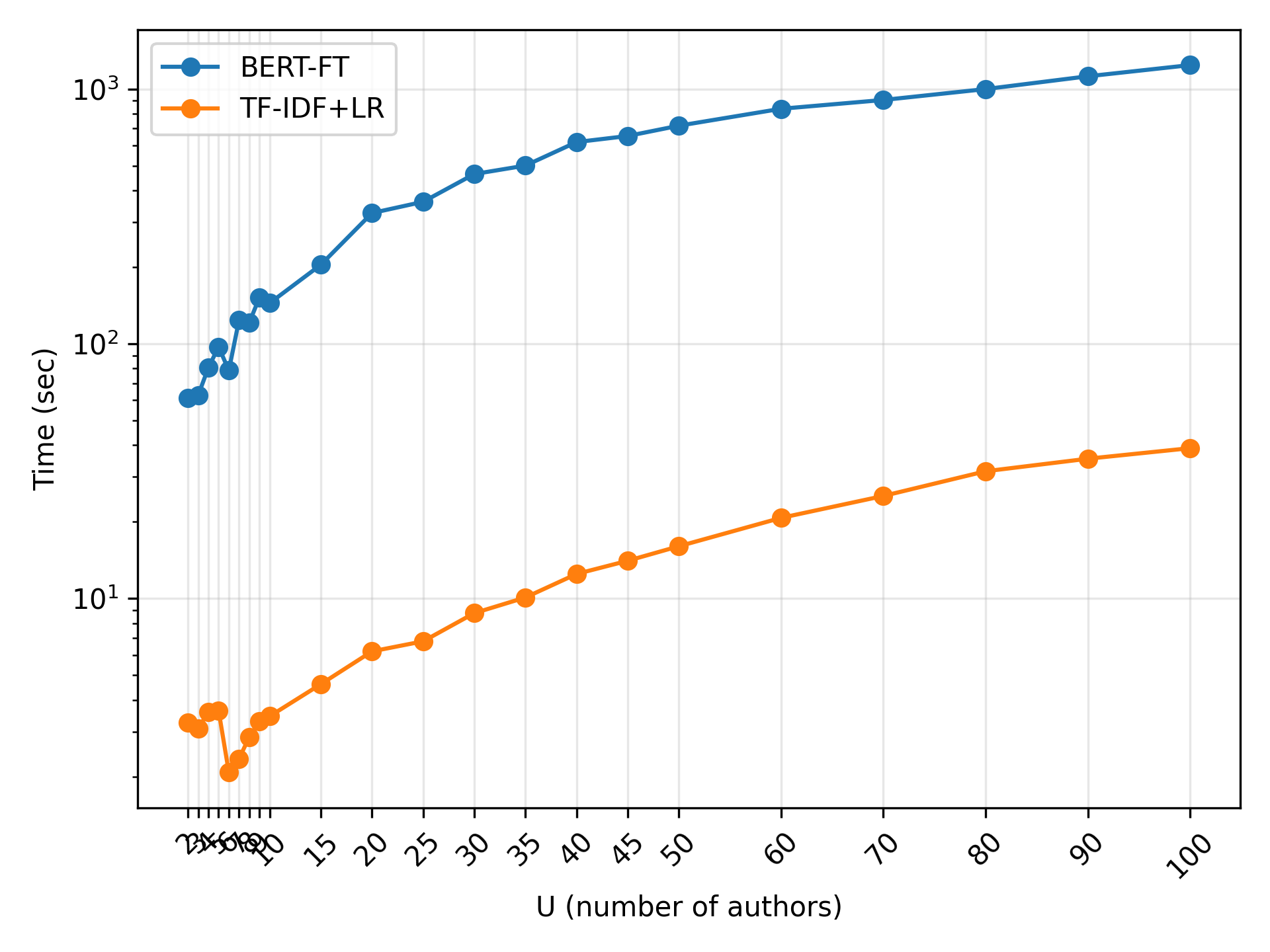}
    \caption{Experiment~3 (small scale): $U=2$--$100$, log scale.}
    \label{fig:exp3_smallU_time}
  \end{minipage}\hfill
  \begin{minipage}{0.49\textwidth}
    \centering
    \includegraphics[width=\linewidth]{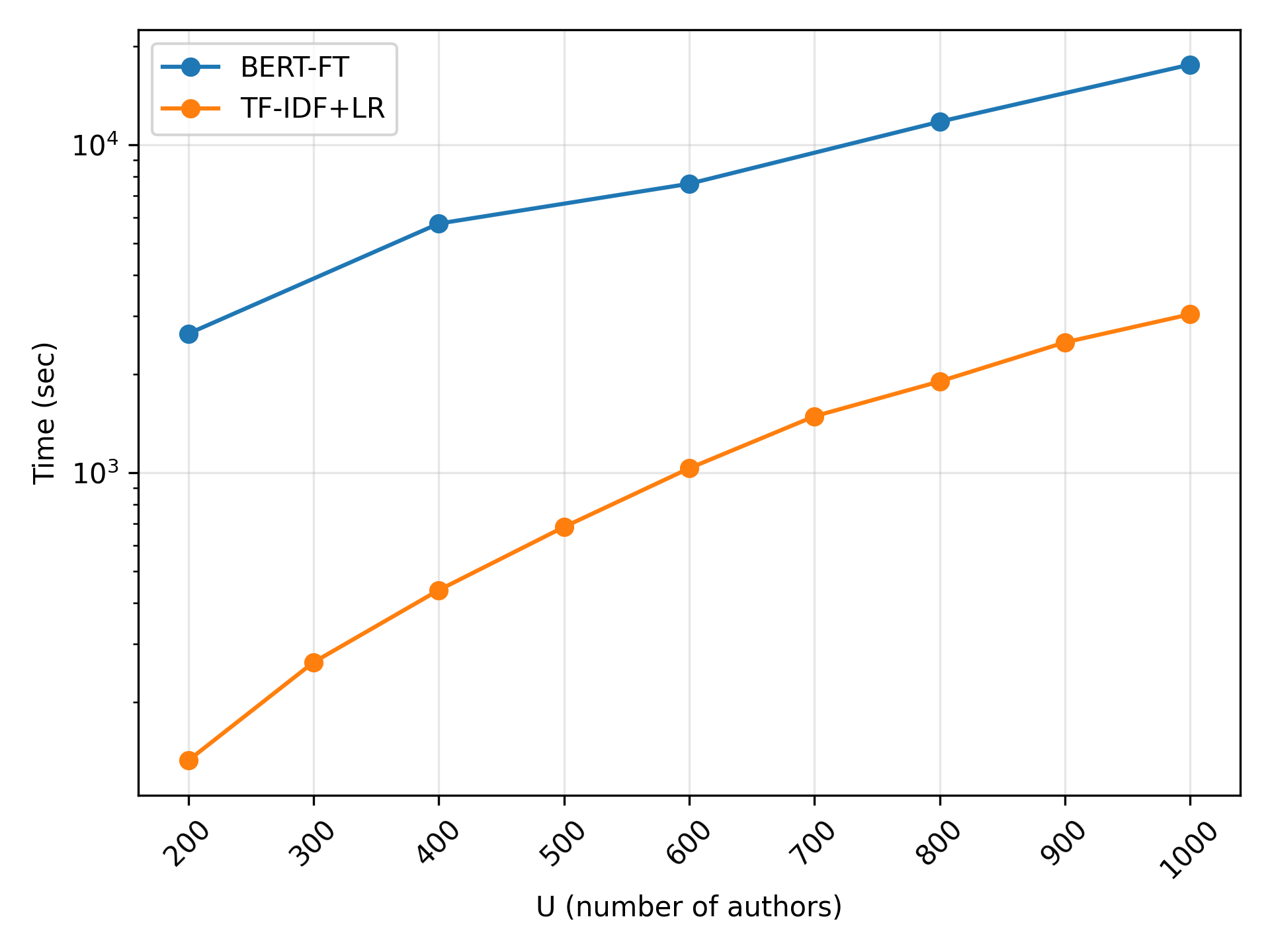}
    \caption{Experiment~3 (large scale): $U=200$--$1{,}000$, log scale.}
    \label{fig:exp3_largeU_time}
  \end{minipage}
\end{figure*}

\noindent\textbf{Results.}
For $U\le100$, both methods maintained Top-10 around 0.95.
At $U=1{,}000$, Top-1 accuracy dropped to approximately 0.51 for both methods, but TF-IDF+LR maintained a Top-10 of 0.74 (Table~\ref{tab:exp3_scale}).
BERT-FT exhibited large standard deviations of 0.17--0.22 around $U=600$--$800$, indicating significant fold-to-fold variability (Figure~\ref{fig:exp3_largeU_macro_f1}).
Training time for TF-IDF+LR increased approximately linearly from 39 seconds at $U=100$ to 3,041 seconds ($\approx$50 minutes) at $U=1{,}000$, while BERT-FT increased more steeply from 1,241 seconds ($\approx$21 minutes) at $U=100$ to 17,564 seconds ($\approx$4.9 hours) at $U=1{,}000$ (Figures~\ref{fig:exp3_smallU_time} and \ref{fig:exp3_largeU_time}).
 
\noindent\textbf{Discussion.}
The instability of BERT-FT is partly attributable to the increasing number of classes, which heightens sensitivity to mixed-precision training (fp16) and learning rate scheduling.
Additionally, as the number of authors grows, the probability of including authors with similar stylistic and lexical profiles increases, expanding the number of decision boundaries and making optimization harder, potentially causing cascading misclassifications in certain folds.
TF-IDF+LR exhibited smooth and predictable performance degradation, demonstrating clear operational advantages in stability and computational cost at large scales.
The fact that Top-1 remains at approximately 0.51 while Top-10 reaches 0.74 at $U=1{,}000$ is meaningful from the practical perspective of reducing 1,000 candidates to 10 (screening).
Note that this experiment was designed based on the number of reviews per author $k$ and did not control for the character count (total token count) of each review.
Since review length varies across authors and conditions, the interpretation of ``how many reviews are sufficient'' may be confounded with ``how much total text is needed.''

\begin{table*}[t]
  \centering
  \scriptsize
  \setlength{\tabcolsep}{3pt}
  \renewcommand{\arraystretch}{1.15}
  \caption{Experiment~3: Scaling characteristics ($k=186$ fixed, 5-fold mean; Acc/Macro-F1 as mean $\pm$ SD, Top-$k$ and time as mean).}
  \label{tab:exp3_scale}
  \begin{tabular}{rrcccccccccc}
    \toprule
    $U$ & $N$ &
    \multicolumn{5}{c}{TF-IDF+LR} & \multicolumn{5}{c}{BERT-FT} \\
    \cmidrule(lr){3-7}\cmidrule(lr){8-12}
     & & Acc & Macro-F1 & Top-5 & Top-10 & Time(s) & Acc & Macro-F1 & Top-5 & Top-10 & Time(s) \\
    \midrule
    2    & 372    & 0.997$\pm$0.005 & 0.997$\pm$0.005 & 1.000 & 1.000 & 3    & 1.000$\pm$0.000 & 1.000$\pm$0.000 & 1.000 & 1.000 & 61 \\
    5    & 930    & 0.975$\pm$0.012 & 0.975$\pm$0.012 & 1.000 & 1.000 & 4    & 0.983$\pm$0.009 & 0.983$\pm$0.009 & 1.000 & 1.000 & 97 \\
    10   & 1,860   & 0.901$\pm$0.012 & 0.900$\pm$0.012 & 0.989 & 1.000 & 3    & 0.936$\pm$0.015 & 0.936$\pm$0.015 & 0.996 & 1.000 & 144 \\
    20   & 3,720   & 0.897$\pm$0.015 & 0.898$\pm$0.015 & 0.978 & 0.994 & 6    & 0.901$\pm$0.014 & 0.901$\pm$0.013 & 0.981 & 0.995 & 326 \\
    50   & 9,300   & 0.867$\pm$0.010 & 0.867$\pm$0.010 & 0.956 & 0.985 & 16   & 0.871$\pm$0.011 & 0.870$\pm$0.011 & 0.967 & 0.988 & 717 \\
    100  & 18,600  & 0.788$\pm$0.005 & 0.788$\pm$0.005 & 0.919 & 0.953 & 39   & 0.792$\pm$0.005 & 0.792$\pm$0.007 & 0.923 & 0.954 & 1,241 \\
    200  & 37,200  & 0.717$\pm$0.005 & 0.713$\pm$0.005 & 0.873 & 0.916 & 132  & 0.727$\pm$0.008 & 0.728$\pm$0.007 & 0.869 & 0.913 & 2,653 \\
    400  & 74,400  & 0.631$\pm$0.004 & 0.622$\pm$0.004 & 0.796 & 0.852 & 438  & 0.642$\pm$0.005 & 0.642$\pm$0.006 & 0.797 & 0.849 & 5,750 \\
    600  & 111,600 & 0.585$\pm$0.003 & 0.571$\pm$0.004 & 0.751 & 0.811 & 1,031 & 0.472$\pm$0.173 & 0.472$\pm$0.173 & 0.609 & 0.661 & 7,604 \\
    800  & 148,800 & 0.545$\pm$0.004 & 0.528$\pm$0.005 & 0.714 & 0.778 & 1,898 & 0.444$\pm$0.221 & 0.443$\pm$0.222 & 0.569 & 0.619 & 11,757 \\
    1000 & 186,000 & 0.510$\pm$0.002 & 0.491$\pm$0.002 & 0.678 & 0.745 & 3,041 & 0.501$\pm$0.025 & 0.499$\pm$0.028 & 0.657 & 0.722 & 17,564 \\
    \bottomrule
  \end{tabular}
\end{table*}

\subsection{Discussion of Misclassification Trends}
\label{subsec:error_discussion}
 
Although this study does not perform a quantitative decomposition of misclassification factors, observation of representative misclassification cases revealed trends consistent with difficulty factors identified in prior work \citep{stamatatos2009survey,stamatatos-2017-authorship,kestemont2018overview}.
 
\subsubsection{Impact of Boilerplate Text (Genre/Register Homogenization)}
 
Reviews frequently contain formulaic expressions such as shipping confirmations and packaging evaluations (e.g., ``Thank you for the prompt response'').
As such expressions increase, inter-author linguistic similarity rises, making style-based discrimination more difficult.
From the perspective of computational stylometry, vocabulary and formulaic expressions derived from genre or topic have been noted to confound identification \citep{stamatatos2009survey}, and this effect is likely amplified in short texts with prevalent boilerplate.
 
\subsubsection{Information Deficiency Due to Short Text (Observation Constraint)}
 
Very short texts such as ``It was good'' or ``Satisfied'' provide insufficient material for author-specific expression to manifest, limiting the cues available to models and increasing misclassification.
The strong influence of the number of authors and per-author data volume on performance has been documented in prior work \citep{luyckx2010effect}, and the short-text nature of reviews constitutes one of the key difficulty factors in this study.
On the other hand, Top-$k$ performance was relatively maintained under certain conditions, suggesting that even when exact identification is difficult, ``narrowing down candidates'' may retain practical utility.
 
\subsubsection{Topic Dependency (Topic Vocabulary Contamination)}
 
When topic-specific vocabulary tied to product categories or product names (e.g., ``smartphone case,'' ``charging speed'') is prevalent, topic vocabulary can contaminate the attribution signal, causing identification based on content rather than style.
This is a well-known challenge in authorship attribution, and various mitigations including topic-neutralization techniques and cross-domain evaluation settings have been explored \citep{stamatatos-2017-authorship,kestemont2018overview}.
Since this study is a foundational evaluation on reviews (single domain), topic dependency may persist.

\section{Conclusion and Future Work}
\label{sec:conclusion}
 
\subsection{Summary}
 
This study used Japanese review data from the clear web to organize the conditions under which authorship attribution can function as candidate screening, examining the effects of the number of authors $U$, posts per author $k$, class imbalance, and computational cost.
 
In method comparison (Experiments~1 and 2), BERT-FT achieved the best performance, but its training cost was approximately 29 times that of TF-IDF+LR.
For large-scale and repeated operation, TF-IDF+LR was confirmed to be a practical choice balancing performance, stability, and computational cost.
In scaling analysis (Experiment~3), both methods maintained practical performance levels for $U\le100$, but BERT-FT training became unstable at scales of several hundred or more, highlighting the advantage of TF-IDF+LR.
 
At $U=100$, BERT-FT achieved Accuracy of 0.77--0.86, while Top-5 reached 0.90--0.95, suggesting that Top-$k$ evaluation supports an operational mode that reduces the risk of overcommitting to a single incorrect author while narrowing down candidates.
For practical deployment, a two-stage architecture---first screening candidates to Top-$k$ with TF-IDF+LR, then refining with BERT-FT---is a viable option that reduces computational cost while pursuing high accuracy, provided that the first stage's $k$ is designed to retain the correct author in the candidate set.

\subsection{Future Work}
\label{subsec:future}
 
\subsubsection{Refinement of Feasibility Conditions (Total Text Volume and Training Settings)}
 
Experiment~1 showed diminishing returns with increasing $k$, and Experiment~3 revealed BERT-FT training instability beyond $U=600$.
These observations suggest that feasibility conditions may be governed not only by the number of posts $k$ but also by total text volume and training settings.
Future work should evaluate performance under conditions where total character count (total token count) is controlled, and expand classifier selection and hyperparameter search for lightweight methods to further improve accuracy.
 
\subsubsection{Addressing Short Text (Collective Attribution)}
 
The data in this study has a median length of 50--60 characters with frequent boilerplate, leading to insufficient author-specific cues and difficulty in discrimination.
Future work will introduce collective attribution, which bundles multiple posts for estimation, and evaluate the degree to which this improves accuracy.
 
\subsubsection{Extension Toward Dark Web Application (Domain Gap and Open-Set)}
 
The findings of this study are based on Japanese, clear web reviews in a single-domain, closed-set setting and may not directly transfer to dark web posts.
While short text, boilerplate, and topic dependency were observed as misclassification factors, dark web posts introduce additional challenges: multilingual content, slang, intentional obfuscation, mixed quotations and template text, topic variability, and multi-account operation, all of which may create more difficult conditions than reviews.
 
Future work should therefore consider:
(i) domain-gap-aware evaluation (cross-domain and domain adaptation);
(ii) open-set identification that handles unknown authors (returning ``no match''); and
(iii) multilingual (e.g., English, Russian) and cross-platform identity linking.

\section*{Acknowledgments}
 
This study used the ``Rakuten Dataset'' provided by Rakuten Group, Inc. via the IDR Dataset Service of the National Institute of Informatics (\url{https://rit.rakuten.com/data_release/}).
We gratefully acknowledge Rakuten Group, Inc. and the National Institute of Informatics for providing this valuable dataset.

\bibliographystyle{unsrtnat}
\bibliography{refs}  






\end{document}